\relax
\documentclass[letterpaper]{article} 
\usepackage{aaai21}  
\usepackage{times}  
\usepackage{helvet} 
\usepackage{courier}  
\usepackage[hyphens]{url}  
\usepackage{graphicx} 
\usepackage{graphicx}  
\usepackage{natbib}  
\usepackage{caption} 

\usepackage{amsmath}
\usepackage{bm}
\usepackage{amsfonts}
\usepackage{amssymb}
\usepackage{amsthm}
\usepackage{bm}
\usepackage{algorithm}
\usepackage{algpseudocode}
\usepackage{color}
\usepackage{multirow}
\usepackage{booktabs}
\title{SSKD: Self-Supervised Knowledge Distillation\\ for Cross Domain Adaptive Person Re-Identification }
\author{
Junhui Yin, Jiayan Qiu, Siqing Zhang, Zhanyu Ma\thanks{Corresponding author}, Jun Guo
}
\affiliations{

}

\begin{document}

\maketitle

\begin{abstract}

Domain adaptive person re-identification (re-ID) is a
challenging task due to the large discrepancy between the source domain and the target domain.
To reduce the domain discrepancy, existing methods mainly attempt to generate pseudo labels for unlabeled target images by clustering algorithms. However, clustering methods tend to bring noisy labels and the rich fine-grained details in unlabeled images are not sufficiently exploited. In this paper, we seek to improve the quality of labels by capturing feature representation from multiple augmented views of unlabeled images. To this end, we propose a Self-Supervised Knowledge Distillation (SSKD) technique containing two modules, the identity learning and the soft label learning. Identity learning explores the relationship between unlabeled samples and predicts their one-hot labels by clustering to give exact information for confidently distinguished images. Soft label learning regards labels as a distribution
and induces an image to be associated with several related
classes for training peer network in a self-supervised manner, where the slowly evolving network is a core to obtain soft labels as a gentle constraint for reliable images. Finally, the two modules can resist label noise for re-ID by enhancing each other and systematically integrating
label information from unlabeled images. Extensive
experiments on several adaptation tasks demonstrate that the proposed method outperforms the current state-of-the-art approaches
by large margins.

\end{abstract}

\section{ Introduction}  

\noindent The motivation of person re-identification (re-ID) is to find a target person's images in one camera from the gallery images obtained by other different cameras. While person re-ID in fully-supervised manner has shown great progress given the recent advances of convolutional neural network \cite{luo2019bag}, person re-ID in unsupervised
domain adaptation (UDA) is still a challenging task. The key problem is that UDA models in person re-ID suffer from domain gap when it transfers knowledge from the labeled source domain to the unlabeled target domain. This domain discrepancy is generally due to the data-bias existing between source data and target data.


\begin{figure}[t]
\centering
\includegraphics[width=1\columnwidth]{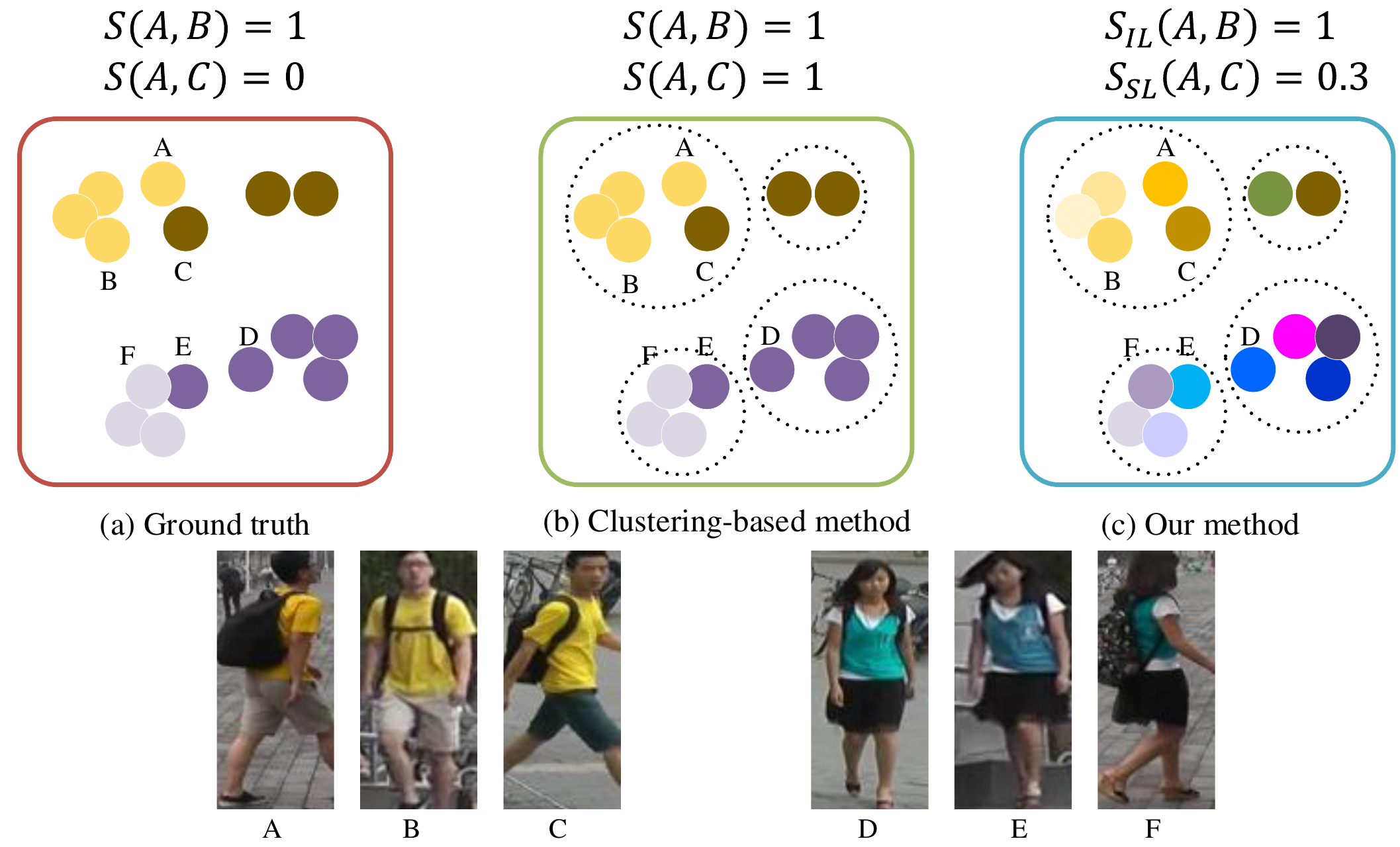}
\caption{(a) The image A and B have the same identity, with an initialized similarity $S(A,B)=1$.
The image C has another identity with the similarity $S(A,C)=0$. Similar results also appear in image D, E, and F. (b) Although image B and C are different persons, the clustering based method roughly divides them into the same cluster due to similar clothes. (c) Our method gives exact information $S_{IL}(A,B)=1$ for confidently distinguished images by identity learning (IL) and a gentle constraint $S_{SL}(A,C)=0.3$ for similar images (colors) by soft label learning (SL).
}
\label{intro}
\end{figure}
Recently, various methods \cite{2018arXiv180711334S,fu2019self,zhang2019self,yang2020asymmetric,ge2020mutual}
have been proposed to alleviate this issue. These works mainly focus on pseudo label generation of unlabeled target domain.
The clustering-based adaptation method \cite{2018arXiv180711334S} first extend existing UDA theories to person re-ID tasks and
employ unsupervised clustering methods (\emph{e.g.}, k-means) to predict pseudo labels for unlabeled target data.
SSG \cite{fu2019self} mines the potential similarity of unlabeled samples to
automatically separate unlabeled data into different clusters, and then each cluster is labeled with pseudo identity. These methods usually treat pseudo labels as ground truth labels to optimize the model, which indicates that their performance highly relies on the clustering accuracy. However, most clustering methods can't guarantee clustering accuracy even with modern approaches. Based on this,
the self-training \cite{zhang2019self} considers different learning
strategies to improve
the quality of labels by exploiting complementary data information. Given that some clustering algorithms (\emph{e.g.}, DBSCAN) directly regard low confidence samples as outliers and give them up during training, ACT \cite{yang2020asymmetric} designs two models to promote each other with noisy pseudo labels. Furthermore, mutual mean-teaching \cite{ge2020mutual} is proposed to reduce the effects of noisy labels in a collaborative training manner, which defines identities by off-line and on-line refined pseudo labels.

Despite promising results achieved by recent works,
the performances of these UDA approaches
is still not satisfactory enough
(\emph{e.g.}, the current best performance on the Market1501 dataset is much lower than its supervised counterparts).
The main reason is that most existing methods focus on the clustering quality
but ignoring the potential information existing in the unlabeled images. During the clustering process, different images of the same identity might be separated into different clusters, which generates the wrong pseudo identities.
One reason of this misclassification is that the pseudo labels are predicted by the knowledge learned from the different domain.
Another reason is that the task itself brings large noise to the data. As shown in the Figure~\ref{intro}, different pedestrians under the same camera may wear similar clothes, while the same person will appear under different cameras for the re-ID task.
To further improve the quality of labels and mine the potential information of unlabeled images, we propose to learn different representations by exploiting differently augmented views (\emph{e.g.}, different crops) of the same image via self-supervised learning \cite{he2020momentum,chen2020simple} in the latent space. More specifically, to extend the framework of the re-ID task for exploiting the potential similarity of unlabeled data, we reformulate our main task as an UDA problem by combining the traditional clustering algorithm with soft label learning, where soft label reflects the image similarity and it can capture obvious similarities among semantic categories
in a smooth way.

\begin{figure*}[t]
\centering
\includegraphics[width=1.9\columnwidth]{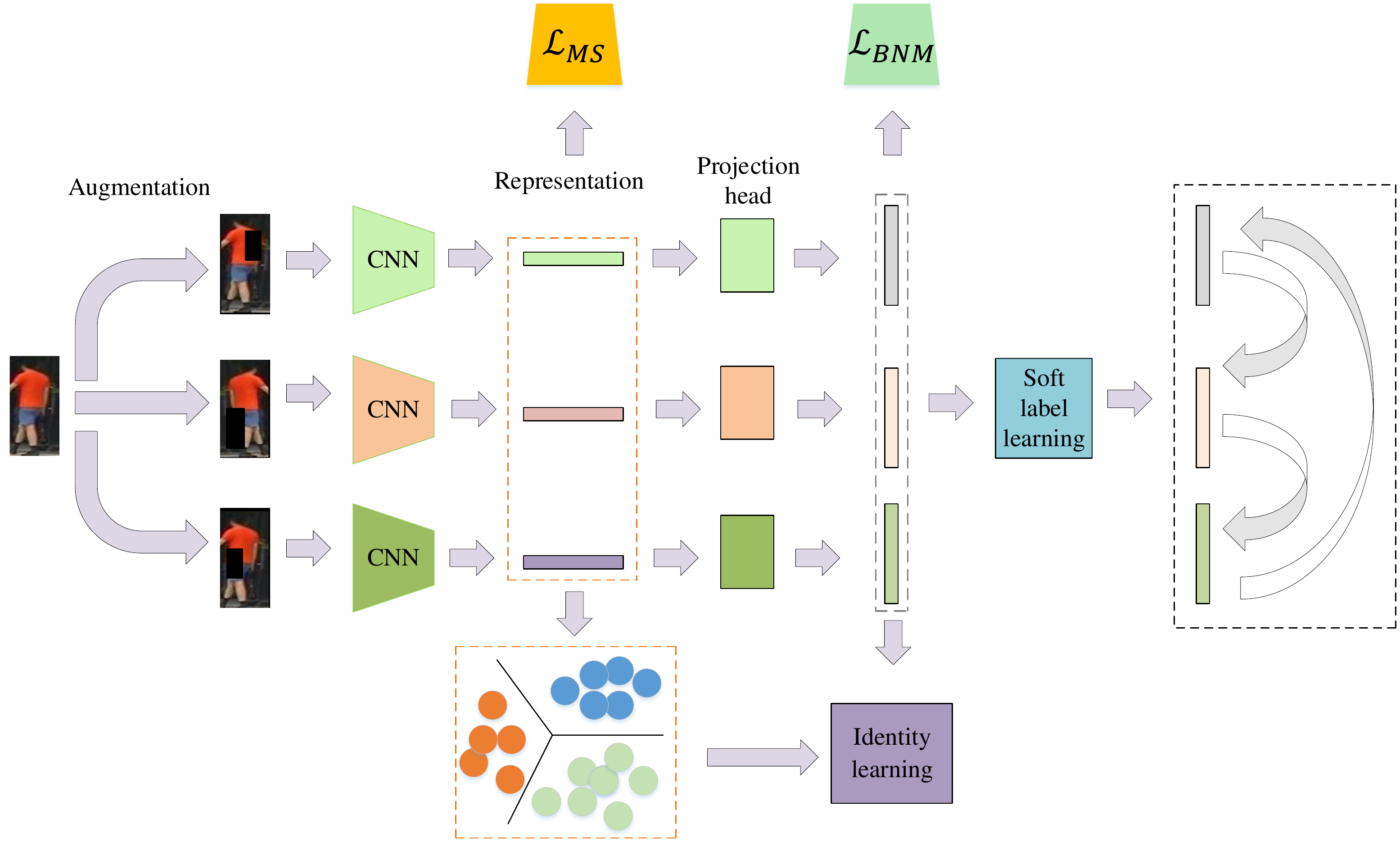}
\caption{The diagram of the proposed self-supervised knowledge distillation (SSKD).
During training, unlabeled images are augmented three times to obtain
 different views of the same images, which are fed-forward into the neural network to
explore multiple similarities between them via $\mathcal{L}_{MS}$.
Subsequently, identity learning and soft label learning are designed to optimize the network with one-hot label and soft label, respectively.
The model further enhances the discriminability and diversity of category prediction by performing $\mathcal{L}_{BNM}$ on batch classification response matrix.
}
\label{fig1}
\end{figure*}

Recently, BNM \cite{cui2020towards} shows progress in boosting model learning capabilities by
reducing ambiguous predictions equipped with
maximizing the batch nuclear-norm on the prediction output matrix, especially in the case of insufficient labels. However, re-ID models in UDA also suffer from insufficient learning scenarios. Therefore, we update BNM to a self-supervised manner to perceive
multiple prediction discriminability and diversity. To this end, we propose Self-Supervised Knowledge Distillation (SSKD), containing two key components:
1) Identity learning, which mines the relationship between unlabeled samples and predicts their one-hot labels by clustering.
2) Soft label learning, which considers labels as a distribution and employs these self-supervised soft labels, obtained from a momentum-based moving average network rather than a self-supervised pretext task, as an auxiliary information to help extracting richer knowledge from the same image's other views. More specifically, identity learning utilizes explicit labels as supervised signals to induce images belonging to an exact class when more confidently labelled samples are incorporated in a progressive way.
Given that domain shift or camera variance
in training images may degrade the CNN feature
representation capability, we introduce mutiple peer models on unseen target domain to distill a powerful model by self-supervised learning.
This self-supervised learning trains a network with soft label, which can be regarded as a gentle constraint to further enhance
the robustness of category prediction against outliers.
Finally, we only keep one network in the inference
stage without additional computing resources or trainable parameters.
Through BNM with identity learning and soft label learning, SSKD can jointly 
motivate an image to be related to several classes for alleviating the noisy identities produced by clustering
and further strengthen the robustness of prediction on categories without extra guidance, as
shown in Figure \ref{fig1}. To support our claims, we evaluate
our approaches on four unsupervised domain adaptation tasks (\emph{e.g.}, Duke-to-Market).
By exploiting unlabeled target dataset with
SSKD, our method outperforms the current state-of-the-art
approaches by large margins.

In summary, our main contributions are three-fold:

$\bullet$ We develop an UDA re-ID framework with progressive
augmentation to combine the identity learning with the soft label learning,
which can induce more confidently labelled images belonging to an exact class
and motivate the low-confident samples to be
associated with several related classes for alleviating the noisy identities.

$\bullet$ We update BNM to a self-supervised manner to better mine the fine-grained characteristics existing in the different augmented views of the same image and enhance the robustness of prediction on categories without access to the target labels.

$\bullet$ SSKD utilizes multiple neural networks that interact and learn from each other. In particular, the slowly evolving networks is a core to obtain self-supervised soft label. In the reference phase, only a powerful model is kept without additional computing resources or trainable parameters. In addition, our approach significantly outperforms the state-of-the-art method by large margins.

\section{ Related Work}

\textbf{Cross domain adaptation person re-ID}. Existing studies in UDA re-ID can be mainly summarized into
two categories of methods. The first category of distribution aligning learns domain-invariant features to overcoming domain discrepancy between different datasets on image-level \cite{deng2018image,wei2018person,bak2018domain} and attribute-level \cite{wang2018transferable,liu2019adaptive,chen2019instance}.
PDA-Net \cite{li2019cross}
further explore CycleGAN \cite{zhu2017unpaired} or StarGAN \cite{choi2018stargan} as a style transformer to improve the discriminative capacity of model by aligning the distribution shift of two domains.
Another line of methods is to exploit the underlying relations among unlabeled images and generate pseudo labels based on clustering \cite{fu2019self,yang2020asymmetric,ge2020mutual}. Recently, AD-Cluster \cite{zhai2020ad} enhances the model
discriminative capacity on unlabeled domain by alternating clustering and sample generation in a min-max optimization scheme. 
However, these works suffer from the problem of the wrong pseudo labels, which produced by clustering and noisy data.

\textbf{Knowledge distillation}. Knowledge distillation is an important technique to learn
a smaller network with better inference efficiency from a larger network with better quality.
\cite{hinton2015distilling} propose to maximize agreement between the output distributions of teacher and student models by minimizing the KL-divergence of these distributions. Many distillation methods utilize the logits of a teacher model as the knowledge \cite{zhang2018deep,xu2020knowledge}, while the intermediate feature maps can also be employed as the knowledge to guide the learning of the student model \cite{ahn2019variational}. In particular, DMT \cite{zhang2018deep} transfers knowledge between
an ensemble of students by collaborative
learning and teach each other. Inspired by DMT, MMT \cite{ge2020mutual} and MEB-Net \cite{zhai2020multiple} improve
the discriminative capability of re-ID models in UDA by integrating multiple sub-networks in a mutual learning manner.
The mutual learning mechanism is performed by soft identity loss and soft softmax triplet loss. However, this learning manner is heavy and need to manually specify the number of identities.
Our SSKD can automatically determine the number of clusters and only employ the logits of network
as supervise signal to transfer knowledge between student models effectively. Finally, our method achieves the state-of-the-art performances
on the several challenge datasets.

\textbf{Learning with noisy data}.
Recently, training network model on noisy data has received extensive attention.
Related approaches can be divided into two categories, which contains robust loss function \cite{ghosh2017robust,zhang2018generalized} and the training
strategies \cite{zhang2019self,yang2020asymmetric}. Some robust loss functions are designed for training model with noisy data, for example,
mean absolute error loss \cite{ghosh2017robust} and generalized cross entropy loss \cite{zhang2018generalized}.
However, these methods could not be directly used for UDA person re-ID.
Some training strategies are proposed for re-ID task. PAST \cite{zhang2019self} is consisted of conservative
stage and promoting stage to select high confidence training samples. ACT \cite{yang2020asymmetric} can not only select possibly clean samples, but also design different data flow for two networks. In this work, we develop two modules, IL and SLL, to resist noisy labels for re-ID task by enhancing each other and systematically integrating label information from unlabeled images.

\section{The Proposed Approach}

Re-ID models have the basic
 discriminative ability for adaptation by training model with labeled source domain.
Our goal is to learn more fine-grained details from unlabeled target images for knowledge transfer between different models, thus we need to first mining potential characteristics. Given a batch of images, each image is augmented three times to create three views of the same sample. Next, we employ the multi-similarity loss \cite{wang2019multi} to
exploit the multiple similarities for the representation features in the latent space.
Furthermore, we maximize agreement between different augmented views of the same image by
self-supervised learning. Finally, we extend BNM to our work for strengthening the
robustness and diversity of prediction on categories in a self-supervised manner.

\textbf{Preliminary}. For unsupervised domain
adaptation (UDA) in person re-ID, we define the source domain data as
$\mathcal{S}=\{(x_{s,i}, y_{s,i})|_{i=1}^{n_{s}}\}$, where $n_{s}$ is the number of person images in the source domain and each image $x_{s,i}$ is associated with an identity label $y_{s,i}$. Similarly, let $\mathcal{T}$ denote a target domain containing $n_{t}$ unlabeled images $\{(x_{t,i})|_{i=1}^{n_{t}}\}$.

\subsection{Supervised Learning for Source Model}
The identity information of source domain are available, thus the training process of the source data can be treated as a conventional classification problem \cite{zheng2016person}.
Source model is optimized by minimizing the cross
entropy loss,
\begin{equation}\label{tik}
\mathcal{L}_{src}= -\frac{1}{n_{s}}\sum_{i=1}^{n_{s}}\log p(y_{s,i}|x_{s,i}),
\end{equation}
where $n_{s}$ is the number of images in source domain. $p(y_{s,i}|x_{s,i})$ is the predicted probability of image $x_{s,i}$
belonging to the identity $y_{s,i}$. At this time, the model has a certain adaptability for cross domain.

\subsection{Self-Supervised Knowledge Distillation}

\textbf{Image augmentation}. To obtain the cross views of feature representation and  prediction results, each image $x_{t,i}$ is augmented three times using random crop, random horizontal flip, and random erasing \cite{zhong2020random}, creating three views of the same image, $\tilde{x}_{t,3i+1}$, $\tilde{x}_{t,3i+2}$ and $\tilde{x}_{t,3i+3}$. The three images are further encoded via three encoder networks with the same architecture, referred to as student models, to generate
representation features $h_{t,3i+1}$, $h_{t,3i+2}$ and $h_{t,3i+3}$.
These feature encoders are driven to perceive
subtle differences of natural characteristic and semantic information
among different augmented views which is essential for the representation of precise
fine-grained details.

\textbf{Multi-similarity learning}.
After obtaining the representation features in the embedding space, we further explore full pair-wise relations between samples in a mini-batch.
Triplet loss \cite{hoffer2015deep} and contrastive loss \cite{hadsell2006dimensionality} are not sufficient to
%
pull similar samples and repel dissimilar ones  for re-ID task. These approaches focus on either cosine similarities
or relative similarities. To exploiting the multiple similarities for the representation features of augmented views of different images, we employ the multi-similarity loss \cite{wang2019multi}, implemented by two iterative steps with sampling
and weighting, for pair weighting in the representation space.
This allows it to compare the representation of an augmented view with other negative samples, providing a more principled method for mining the fine-grained details existing in the differently augmented views of samples.

Let the similarity of two features be
$S_{ij}=<h_{t,i},h_{t,j}>$, where $<\cdot,\cdot>$ defines
dot product, yielding a similarity matrix $\bm{S}$ whose element
at $(i,j)$ is $S_{ij}$.
Multi-similarity (MS) loss is formulated as,
\begin{equation}
\begin{aligned}\label{MS}
\mathcal{L}_{MS}=&-\frac{1}{n_{t}}\sum_{i=1}^{n_{t}}\{\frac{1}{\alpha}\log[1+\sum_{k\in\mathcal{P}_{i}}e^{-\alpha(S_{ik-\lambda})}]\\
&+\frac{1}{\beta}\log[1+\sum_{k\in\mathcal{N}_{i}}e^{\beta(S_{ik}-\lambda)}]\},
\end{aligned}
\end{equation}
where $n_{t}$ is the number of images in target domain, and $\alpha, \beta$ are hyper-parameters as in \cite{wang2019multi}

\textbf{Identity learning}.
Aiming at aggregating the information from different augmented views of the same image, we design a
fusion operation to generate an individual class for each training image.
To learn richer representations, the operation should
effectively confuse different features, while reserving the distinctive information.
Under this principle, we sum three features and then average them, formulated as,
\begin{equation}
\begin{aligned}
h_{t,i}=\frac{1}{3}\sum_{j=1}^{3}h_{t,3i+j}, i=1,2,\cdots,n_{t}.
\end{aligned}
\end{equation}
Based on these fused features,
we perform clustering algorithm \cite{ester1996density} on them,
which leads to every person image assigned
with a pseudo label according to clustering results. As a result, we can establish
the pseudo label for each image and the number of person identities, denoted as $\hat{y}_{t,i}$ and $m_{t}$ respectively.
The $k$-$th$ network
model $f(\cdot|\theta_{k})$ is updated by optimizing a cross entropy loss with label smoothing, defined as
\begin{equation}
\begin{aligned}
\mathcal{L}_{id}^{k}=\frac{1}{n_{t}}\sum_{i=1}^{n_{t}}\sum_{j=1}^{m_{t}}q_{j}\log p_{j}(x_{t,i}|\theta_{k}),
\end{aligned}
\end{equation}
where $q_{j}=1-\varepsilon+\frac{\epsilon}{m_{t}}$ if $q_{j}=\hat{y}_{t,i}$, otherwise $q_{j}=\frac{\varepsilon}{m_{t}}$.
The overall loss for identity learning can be expressed as
$\mathcal{L}_{id}=\sum_{k=1}^{3}\mathcal{L}_{id}^{k}$.
Comparing with cross entropy loss, $\mathcal{L}_{id}$ replace the original label distribution
with a mixture of the label distribution and the uniform distribution, leading to label-smoothing regularization.

\begin{figure}[t]
\centering
\includegraphics[width=1\columnwidth]{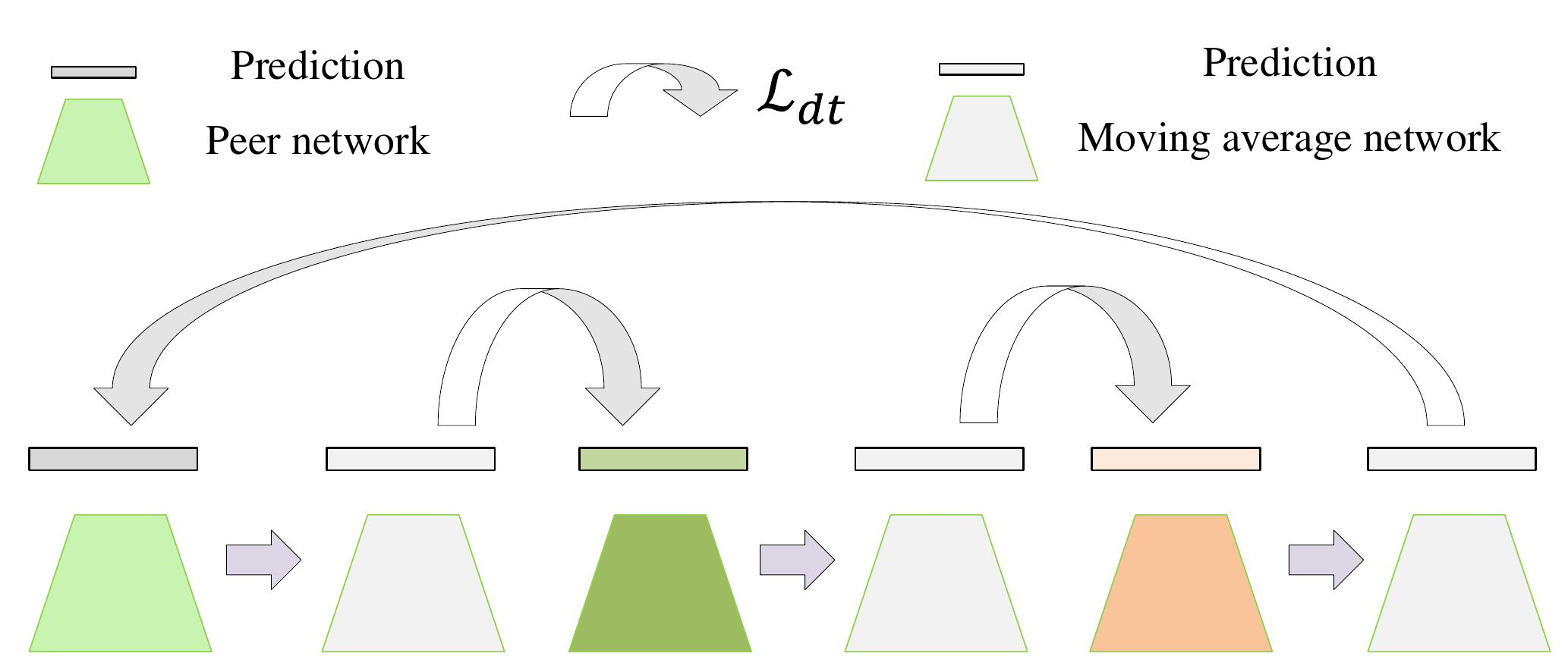}
\caption{The momentum update mechanism for transferring knowledge from a peer network to another network in a self-supervised manner.}
\label{momentum update}
\end{figure}

\begin{table*}[htbp]
  \centering
  \caption{Comparison with state-of-the-art methods on Dukemtmc-reID (Duke) and Market-1501 (Market) for domain adaptive tasks.}
    \begin{tabular}{cccccc|cccc}
    \toprule
    \multicolumn{2}{c|}{\multirow{2}[4]{*}{Methods}} & \multicolumn{4}{c|}{Duke $\rightarrow$ Market} & \multicolumn{4}{c}{Market $\rightarrow$ Duke} \\
\cmidrule{3-10}    \multicolumn{2}{c|}{} & \multicolumn{1}{c|}{mAP} & \multicolumn{1}{c|}{top-1} & \multicolumn{1}{c|}{top-5} & top-10 & \multicolumn{1}{c|}{mAP} & \multicolumn{1}{c|}{top-1} & \multicolumn{1}{c|}{top-5} & top-10 \\
    \midrule
    \midrule
    \multicolumn{1}{l|}{PUL} & \multicolumn{1}{l|}{TOMM'18} & 20.5  & 45.5  & 60.7  & 66.7  & 16.4  & 30.0    & 43.4  & 48.5 \\
    \multicolumn{1}{l|}{TJ-AIDL} & \multicolumn{1}{l|}{CVPR'18} & 26.5  & 58.2  & 74.8  & 81.1  & 23.0    & 44.3  & 59.6  & 65.0 \\
    \multicolumn{1}{l|}{SPGAN} & \multicolumn{1}{l|}{CVPR'18} & 22.8  & 51.5  & 70.1  & 76.8  & 22.3  & 41.1  & 56.6  & 63.0 \\
    \multicolumn{1}{l|}{HHL} & \multicolumn{1}{l|}{ECCV'18} & 31.4  & 62.2  & 78.8  & 84.0    & 27.2  & 46.9  & 61.0    & 66.7 \\
    \multicolumn{1}{l|}{UCDA} & \multicolumn{1}{l|}{ICCV'19} & 30.9  & 60.4  &   -    &   -    & 31.0    & 47.7  &   -    &  - \\
    \multicolumn{1}{l|}{PDA-Net} & \multicolumn{1}{l|}{ICCV'19} & 47.6  & 75.2  & 86.3  & 90.2  & 45.1  & 63.2  & 77.0    & 82.5 \\
    \multicolumn{1}{l|}{CR-GAN} & \multicolumn{1}{l|}{ICCV'19} & 54.0    & 77.7  & 89.7  & 92.7  & 48.6  & 68.9  & 80.2  & 84.7 \\
    \multicolumn{1}{l|}{PCB-PAST} & \multicolumn{1}{l|}{ICCV'19} & 54.6  & 78.4  &    -   &     -  & 54.3  & 72.4  &  -     &  \\
    \multicolumn{1}{l|}{SSG} & \multicolumn{1}{l|}{ICCV'19} & 58.3  & 80.0    & 90.0    & 92.4  & 53.4  & 73.0    & 80.6  & 83.2 \\
    \multicolumn{1}{l|}{ECN++} & \multicolumn{1}{l|}{TPAMI'20} & 63.8  & 84.1  & 92.8  & 95.4  & 54.4  & 74.0    & 83.7  & 87.4 \\
    \multicolumn{1}{l|}{MMCL} & \multicolumn{1}{l|}{CVPR'20} & 60.4  & 84.4  & 92.8  & 95.0    & 51.4  & 72.4  &  82.9  & 85.0 \\
    \multicolumn{1}{l|}{SNR} & \multicolumn{1}{l|}{CVPR'20} & 61.7  & 82.8  &    -  &  -   & 58.1  & 76.3  &  -   & -  \\
    \multicolumn{1}{l|}{AD-Cluster} & \multicolumn{1}{l|}{CVPR'20} & 68.3  & 86.7  & 94.4  & 96.5  & 54.1  & 72.6  & 82.5  & 85.5 \\
    \multicolumn{1}{l|}{MMT} & \multicolumn{1}{l|}{ICLR'20} & 71.2  & 87.7  & 94.9  & 96.9  & 65.1  & 78.0    & 88.8  & 92.5 \\
    \multicolumn{1}{l|}{DG-Net++} & \multicolumn{1}{l|}{ECCV'20} & 61.7  & 82.1  & 90.2  & 92.7   & 63.8 &  78.9 &  87.8&  90.4   \\
    \multicolumn{1}{l|}{MEB-Net} & \multicolumn{1}{l|}{ECCV'20} & 76.0  & 89.9  & 96.0  & 97.5  & 66.1  & 79.6    & 88.3  & 92.2 \\
    \midrule
    \multicolumn{2}{c|}{Ours} & \textbf{78.7} & \textbf{91.7} & \textbf{97.2} & \textbf{98.2} & \textbf{67.2} & \textbf{80.2} & \textbf{90.6} & \textbf{93.3} \\
    \bottomrule
    \end{tabular}%
  \label{tab:duke and market}%
\end{table*}%

\textbf{Self-supervised Learning}.
Due to the domain gap, the pseudo identities generated by
clustering suffer from noises. Instead of the clustering-based method treating reliable samples as the same cluster for
training, we propose a soft label learning, a complementary part of identity learning, that captures obvious similarities
among semantic categories in a smooth way.
It is important to note that the soft similarity is not learned from explicit guidance, but from the visual data itself.
Specifically, the encoded representations are fed into a linear
transformation network (a projection head), resulting in the soft
pseudo labels $z_{t,3i+k}$ for data $x_{t,i}$ with the $k$-$th$ encoder network.
The proposed learning procedure can be regarded as discriminative perception
based on a probability $p^{c}_{3i+k}$ of which the input
image belongs to class $c, p^{c}_{3i+k}=\frac{\exp(z_{t,3i+k}^{c})}{\sum_{c=1}^{C}\exp(z_{t,3i+k}^{c})}$,
where $z_{t,3i+k}^{c}$
is the $c$-$th$ output of the $k$-$th$ projection head.

$Momentum$ $update$. When the above
class predictions are directly served as soft labels to supervise another peer network,
we obtain poor results in experiments. Such case
is caused by the rapidly changing encoder that produces an
error amplification. To enhance the
model predictions' consistency, we propose a momentum-based moving average model for each peer network.
Formally, let the parameter of encoder network $f_{k}$ be $\theta_{k}$. The temporal average model is updated by
\begin{equation}
\begin{aligned}
\theta_{k,e}^{(t)}\leftarrow\rho\theta_{k,e}^{(t-1)}+(1-\rho)\theta_{k},
\end{aligned}
\end{equation}
where $\rho\in [0,1)$ is a momentum coefficient, $t$ is the current iteration, and the initial temporal average parameters $\theta_{k,e}^{(0)}$ are $\theta_{k}$. 

Next, we employ the one network's past evolving model to generate supervised signals for training the other network.
As shown in Figure \ref{momentum update},
the soft label supervision are generated by the three slowly evolving models as $p_{1}(f(x_{3i+1}|\theta_{1,e})), p_{2}(f(x_{3i+2}|\theta_{2,e})),$ and $ p_{3}(f(x_{3i+3}|\theta_{3,e}))$ respectively. The distillation loss with soft label for each network
are defined as
\begin{equation}
\begin{aligned}
&\mathcal{L}_{dt}^{1}(\theta_{1}|\theta_{3})=\frac{1}{n_{t}}\sum_{i=1}^{n_{t}}p_{3}(f(x_{3i+3}|\theta_{3,e}))\log p_{1}(f(x_{3i+1}|\theta_{1})),
\\
&\mathcal{L}_{dt}^{2}(\theta_{2}|\theta_{1})=\frac{1}{n_{t}}\sum_{i=1}^{n_{t}}p_{1}(f(x_{3i+1}|\theta_{1,e}))\log p_{2}(f(x_{3i+2}|\theta_{2})),
\\
&\mathcal{L}_{dt}^{3}(\theta_{3}|\theta_{2})=\frac{1}{n_{t}}\sum_{i=1}^{n_{t}}p_{2}(f(x_{3i+2}|\theta_{2,e}))\log p_{3}(f(x_{3i+3}|\theta_{3})).
\end{aligned}
\end{equation}

The overall distillation loss is
\begin{equation}
\begin{aligned}
\mathcal{L}_{dt}=\mathcal{L}_{dt}^{1}(\theta_{1}|\theta_{3})+\mathcal{L}_{dt}^{2}(\theta_{2}|\theta_{1})
+\mathcal{L}_{dt}^{3}(\theta_{3}|\theta_{2}).
\end{aligned}
\end{equation}

\textbf{Prediction discriminability and diversity}. During training, we want the network could not only predict
training samples into reliable classes, but make it acceptable
to have a certain ability of prediction discriminability and
diversity. BNM \cite{cui2020towards} is proposed to boost the model learning capability under label
insufficient scenarios by reducing the ambiguous
predictions. Due to its simplicity and effectiveness, BNM is adopted in our work to enhance the discrimination capacity of model.
Formally, given a mini batch data with $B$
randomly selected unlabeled images, we define the number of classes as $C$ and introduce the batch prediction output matrix as $A\in\mathbb{R}^{B\times C}$ whose element at $(i,j)$ is $a_{ij}$,
\begin{equation}
\begin{aligned}
\sum_{j=1}^{C}a_{ij}&=1, \forall i \in 1,\cdots,B,
\\
a_{ij}\geq 0,&  \forall i \in 1,\cdots,B,j\in 1,\cdots,C.
\end{aligned}
\end{equation}
 After taking model predictions as pseudo identities, we
further strengthen the robustness of prediction on categories without any extra guidance, using the nuclear-norm maximization of batch classification
response matrix.
Actually, stronger robustness means less uncertainty
in the prediction. This achieved by optimizing the following loss of BNM.
\begin{equation}
\begin{aligned}
\mathcal{L}_{BNM}=\sum_{k=1}^{3}\|A_{k}\|_{*}+\sum_{k=1}^{3}\|A_{k,e}\|_{*}.
\end{aligned}
\end{equation}
Here $A_{k}$ and $A_{k,e}$ stand for the $k$-$th$ network model and its past evolving model respectively.
$\|\cdot\|_{*}$ presents the the calculation of nuclear-norm, which is referred as \cite{cui2020towards}.

$Overall$ $loss$.
By adding the cross entropy loss $\mathcal{L}_{id}^{k}$, the distillation loss $\mathcal{L}_{dt}$, and
the BNM loss $\mathcal{L}_{BNM}$ into Eq. (2), the complete
loss of our final proposed Self-Supervised Knowledge Distillation (SSKD) can be expressed as
\begin{equation}
\begin{aligned}
\mathcal{L}=\xi\mathcal{L}_{id}+(1-\xi)\mathcal{L}_{dt}+\lambda\mathcal{L}_{BNM}+\eta\mathcal{L}_{MS},
\end{aligned}
\end{equation}
where $\xi\in(0,1)$ controls the importance of IL and SSL, and $\lambda, \eta$ are weighting parameters.
The main training process are shown
in Algorithm 1, which can be found in supplementary.

\section{Experimental results and discussions}

\subsection{Experiment setting}
\textbf{Datasets and evaluation metrics}.
We conduct experiments on three common
datasets: Market-1501 \cite{zheng2015scalable}, DukeMTMC-reID \cite{ristani2016performance}, and MSMT17 \cite{ristani2016performance}.
The proposed method is evaluated on four cross domain person Re-ID tasks, which contains Duke-to-Market, Market-to-Duke, Duke-to-MSMT, and Market-to-MSMT. The mean Average Precision (mAP) and Cumulative
Matching Characteristic (CMC) curve are
adopted as the evaluation metrics.

\textbf{Implementation details}.
Random cropping, flipping
and random erasing \cite{zhong2020random} are employed
as the data augmentation. We adopt ResNet-50 \cite{he2016deep} as the backbone of
our model with the last classification layer removed, and initialize it by the ImageNet \cite{deng2009imagenet} pretrained
model.
DBSCAN \cite{ester1996density} is used for clustering before each epoch.
We choose the hyper-parameters (\emph{e.g.}, $\xi,\lambda,\eta$) according to the validation set of
the Duke-to-Market adaptation and directly apply them to other three cross domain re-ID tasks.
For more experiment details, please refer to the supplementary.

\subsection{Comparison with state-of-the-arts}

We compare our proposed approach with existing
techniques on three benchmarks and give the results in Table 1 and Table 2.
Our models achieve state-of-the-art (SOTA) performance in several adaptation tasks.

\textbf{Dukemtmc-reID and Market-1501}. Table \ref{tab:duke and market} reports the quantitative
evaluation results on the Dukemtmc-reID and Market-1501.
Our SSKD achieves an mAP of 78.7\% and a top-1 accuracy of 91.7\% for the
Duke-to-Market, which surpasses the state-of-the-art method (MEB-Net)
 by large margins (2.7\% for mAP and 1.8\% for top-1).
Similar results can be observed in another adaptation task.
The proposed method obtains an mAP of 67.2\% and a top-1 accuracy of 80.2\% for Market-to-Duke, and achieves new state-of-the-art accuracy compared with
the current techniques.
Moreover, our method does not explore how to manually specify the number of
identities in the training set of unlabeled target data, comparing with the current best methods (MMT and MEB-
Net).

\begin{table}[htbp]
  \centering
  \caption{Comparison of proposed method with state-of-art
unsupervised domain adaptive person re-ID methods on MSMT17
dataset.}
    \begin{tabular}{cc|cccc}
    \toprule
    \multicolumn{2}{c|}{Duke $\rightarrow$MSMT} & \multicolumn{1}{c}{mAP} & \multicolumn{1}{c}{top-1} & \multicolumn{1}{c}{top-5} & top-10 \\
    \midrule
    \midrule
    \multicolumn{1}{l|}{PT-GAN} & \multicolumn{1}{l|}{CVPR'18} & 3.3   & 11.8  &   -    & 27.4   \\
    \multicolumn{1}{l|}{SSG} & \multicolumn{1}{l|}{ICCV'19} & 13.3  & 32.2  &    -   & 51.2   \\
    \multicolumn{1}{l|}{ECN++} & \multicolumn{1}{l|}{TPAMI'20} & 16.0    & 42.5  & 55.9  & 61.5 \\
    \multicolumn{1}{l|}{MMCL} & \multicolumn{1}{l|}{CVPR'20} & 16.2  & 43.6  & 54.3  & 58.9 \\
    \multicolumn{1}{l|}{DG-Net++} & \multicolumn{1}{l|}{ECCV'20}  & 22.1  &48.8 & 60.9 & 65.9    \\
    \multicolumn{1}{l|}{MMT} & \multicolumn{1}{l|}{ICLR'20} & 23.3  & 50.1  & 63.9  & 69.8   \\
    \midrule
    \multicolumn{2}{c|}{Ours} & \textbf{26.0} & \textbf{53.8} & \textbf{66.6} & \textbf{72.0} \\
    \bottomrule

    \multicolumn{2}{c|}{Market $\rightarrow$MSMT} & \multicolumn{1}{c}{mAP} & \multicolumn{1}{c}{top-1} & \multicolumn{1}{c}{top-5} & top-10 \\
    \midrule
    \midrule
    \multicolumn{1}{l|}{PT-GAN} & \multicolumn{1}{l|}{CVPR'18} & 2.9   & 10.2  &     -  & 24.4   \\
    \multicolumn{1}{l|}{SSG} & \multicolumn{1}{l|}{ICCV'19} &  13.2  & 31.6  &   -    & 49.6   \\
    \multicolumn{1}{l|}{ECN++} & \multicolumn{1}{l|}{TPAMI'20} & 15.2  & 40.4  & 53.1  & 58.7 \\
    \multicolumn{1}{l|}{MMCL} & \multicolumn{1}{l|}{CVPR'20} & 15.1  & 40.8  & 51.8  & 56.7 \\
    \multicolumn{1}{l|}{DG-Net++} & \multicolumn{1}{l|}{ECCV'20} &22.1 & 48.4 & 60.9 & 66.1    \\
    \multicolumn{1}{l|}{MMT} & \multicolumn{1}{l|}{ICLR'20} & 22.9  & 49.2  & \textbf{63.1} & \textbf{68.8}  \\
    \midrule
    \multicolumn{2}{c|}{Ours} & \textbf{23.8} & \textbf{49.6} & \textbf{63.1} & \textbf{68.8} \\
    \bottomrule
    \end{tabular}%
  \label{tab:MSMT17}%
\end{table}%

\textbf{MSMT17}. Our proposed method is also evaluated on a larger and more challenging
dataset (MSMT17). From experimental results in Table \ref{tab:MSMT17}, it can be observed that
the proposed method
clearly outperforms other methods when DukeMTMC-reID
and Market-1501 are employed as source domains. For instance, with DukeMTMC-reID as source domain,
our approach improves the
accuracy by more than 2.5 percentage points over
the state-of-the-art accuracy of mAP. Our method has better performance than other existing techniques
on all experiments, which further
demonstrates the validation of our proposed method.
More importantly, our method almost approaches the performances of supervised learning \cite{luo2019bag} without any auxiliary information on the target domain.

\begin{figure*}[t]
\centering
\includegraphics[height=0.47\columnwidth]{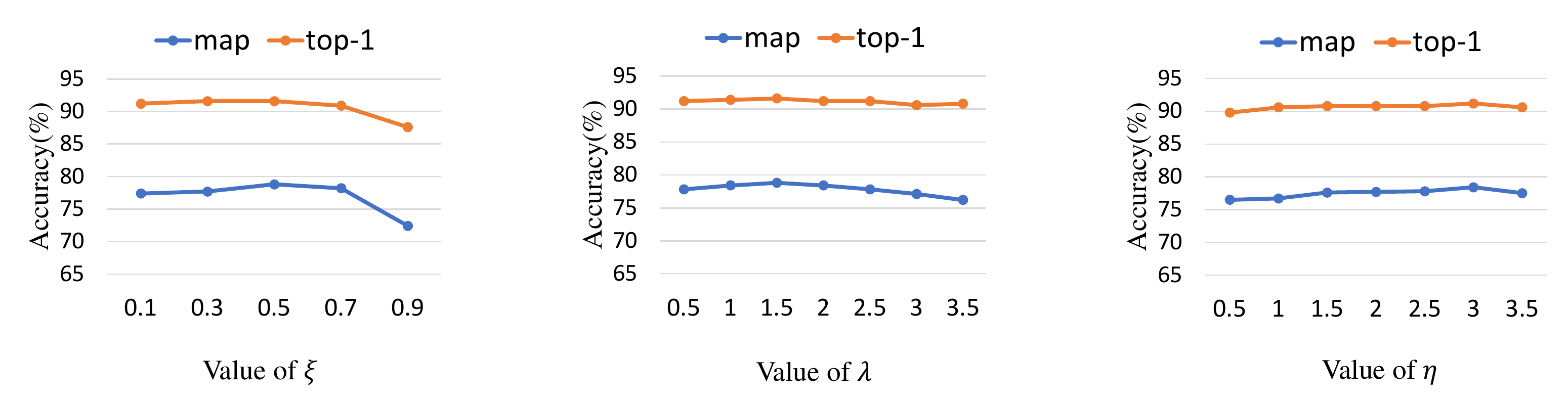}
\caption{Performances of our framework with different values of parameters $\xi$, $\lambda$, and $\eta$.}
\label{para}
\end{figure*}

\subsection{Ablation Study}

Our proposed framework achieves new SOTA performance by using
SSKD. To claim the effectiveness of each component in our framework,
extensive ablation experiments are conducted under different settings.

\textbf{Identity learning}. We propose the identity learning (IL) to predict the ground truth classes for high-confident samples, which
preserve the most
reliable clusters for providing stronger supervision. To verify the effectiveness of such a
learning manner, we evaluate our framework when removing it.
The considerable degrades (\emph{e.g.}, from 78.7\% to 51.3\% for mAP in
Table \ref{tab: Ablation}) are observed under this setting (SSKD (w/o $\mathcal{L}_{id}$)),
especially for Duke-to-Market adaptation. This indicates the importance of adopting the identity learning.

\textbf{Soft label learning}. As a complementary component of identity learning,  soft label learning (SLL) can regard labels as a distribution and enhance the robustness of category prediction. As in Table \ref{tab: Ablation}, on Duke-to-Market task, our SSKD
beats SSKD (w/o $\mathcal{L}_{dt}$) by 78.7 and 91.7 percentage
points on mAP and top-1 respectively.
Similarly, for Market-to-Duke, obvious drops on mAP and top-1 are 1.5\% and
1.6\%, respectively.
The large
performance gaps demonstrate
the effectiveness of the SLL module. Without SLL,
images of the same identity from noisy samples are hard to be
selected as reliable images because of domain shift and camera variance.
In addition, invariance learning has an important effect in this training process by using the momentum update mechanism.
We also demonstrates the importance of adopting the momentum update mechanism, which can be referred in the supplementary.

\textbf{Multi-similarity learning}.
To investigate the impact of multi-similarity loss, we evaluate the performance of our method without multi-similarity learning (MSL).
This modification brings the distinct drops of the performance, \emph{e.g.}, with an over 3\% decrease
of mAP on Market-to-Duke. MSL allows us to mining the fine-grained details
existing in the differently augmented views of the same image. It can integrate the three similarities effectively into a
single framework by using only one loss function.


\textbf{Insufficient learning scenarios}. Our learning task is an unsupervised
open-set domain adaptation, where some labels are insufficient and even unseen.
The methods without the
BNM loss result in lower performances. We could see the significant progress and
benifit of BNM to our re-ID tasks.

\begin{table}[htbp]
  \centering
  \caption{Ablation studies of the proposed approach on individual components for Duke-to-Market and Market-to-Duke adaptation tasks.
 The more results on other metrics can be referred in the supplementary.}
    \begin{tabular}{ccc|cc}
    \toprule
\multicolumn{1}{c|}{\multirow{2}[4]{*}{Methods}} & \multicolumn{2}{c|}{Duke$\rightarrow$ Market} & \multicolumn{2}{c}{Market$\rightarrow$Duke }  \\
\cmidrule{2-5}    \multicolumn{1}{c|}{} & \multicolumn{1}{c}{mAP} & \multicolumn{1}{c|}{top-1} & \multicolumn{1}{c}{mAP} & top-1  \\
    \midrule
    \midrule
    \multicolumn{1}{l|}{SSKD (w/o $\mathcal{L}_{id}$)}  & 51.3  & 74.5  & 50.2  & 67.1    \\
    \multicolumn{1}{l|}{SSKD (w/o $\mathcal{L}_{dt}$)}  & 77.4  & 91.2  & 65.7    & 78.6  \\
    \multicolumn{1}{l|}{SSKD (w/o $\mathcal{L}_{BNM}$)}  & 77.8  & 90.6  & 66.4  & 79.7   \\
    \multicolumn{1}{l|}{SSKD (w/o $\mathcal{L}_{MS}$)}  & 75.9  & 89.8   & 64.0  & 77.5   \\
    \midrule
    \multicolumn{1}{c|}{SSKD} & 78.7 & 91.7 &67.2 & 80.2  \\


    \bottomrule
    \end{tabular}%
  \label{tab: Ablation}%
\end{table}%

\subsection{Parameter Analysis}

To analyze the
effect of different hyper-parameters on adaptation re-ID task, we conduct some comparative experiments by changing
the value of one parameter and keep the others fixed.
We choose the hyper-parameters on
Duke-to-Market task and directly apply them to other
three adaptation tasks.

\textbf{The impact of the hyper-parameter $\bm{\xi}$}.
We first investigate the impact of the hyper-parameter $\xi$ that balances the effect of
the explicit classes and the reliable classes. As shown in Figure \ref{para},
when $\xi$ approaches 0 and 1, the model performance continue decreasing.
At this time, the model reduces to IL and SLL respectively. The model achieves optimal performances when
$\xi=0.5$. The reason is that the weight factor better balances two modules
and induces model to predict ground truth classes for clean label and soft
similarity for noisy label in a progressive way.

\textbf{The impact of the hyper-parameter $\bm{\lambda}$ and $\bm{\eta}$}.
Figure
\ref{para} shows how the re-ID performance varies with different values of $\lambda$ and $\eta$,
weight factors for BNM and MS losses. 
When $\lambda$ increases from 0.5 to 1.5, the re-ID
performance has a slight
increase. When $\lambda$ continually gets
larger, we observe a slight
decrease on re-ID performance. The weight factor $\eta$ also has the similar result, which further indicates that our method is robust and insensitive
to different parameters. Finally, the proposed framework obtains the best result with $\xi=0.5$, $\lambda=1.5$, and $\eta=3$ on
Duke-to-Market transfer.

\section{Conclusions}

In this work, we introduce a novel approach to effectively exploit the potential similarity of unlabeled data to boost performance for cross domain person re-ID task without target labels.
To alleviate the problem of noisy labels produced by clustering, we propose \textbf{Self-Supervised Knowledge Distillation (SSKD)} to jointly motivate an image to be related to several classes and enhance the robustness of category prediction by two key components: $identity$ $learning$ and $soft$ $label$ $learning$. The experimental results indicate that the proposed method achieves new SOTA performances on three challenging datasets.
In addition, we do not increase additional computing resources or trainable parameters in inference stage.

\section{Appendix}

In the supplementary material, we would like to show
more details about implementation detail and experimental results.

\subsection{Experiment detail}
Random cropping, flipping
and random erasing \cite{zhong2020random} are employed
as the data augmentation. We adopt ResNet-50 \cite{he2016deep} as the backbone of
our model with the last classification layer removed, and initialize it by the ImageNet \cite{deng2009imagenet} pretrained
model.
DBSCAN \cite{ester1996density} is used for clustering before each epoch.
We choose the hyper-parameters (\emph{e.g.}, $\xi,\lambda,\eta$) according to the validation set of
the Duke-to-Market adaptation and directly apply them to other three cross domain re-ID tasks.
Specifically, the overall learning process can be trained by two stages: supervised learning for source model and domain adaptation stage.
The main training process is shown
in Algorithm 1.

\textbf{Supervised learning for source model}.
We first pre-train three source models on the source dataset in a supervised manner. These three models has the same architecture: ResNet-50,
and are initialized by using parameters pre-trained on the ImageNet.
The networks parameters $\theta_{1}, \theta_{2},$ and $\theta_{3}$ are optimized independently by minimizing loss. ADAM is employed as our optimizer.
We also utilize a mini-batch size of 64 in 4
GPUs, and set an initial learning rate as 0.00035. Finally, 80
epochs are trained with the learning rate multiplied by 0.1 at 40 and
70 epochs.

\textbf{Cross domain adaptation}.
Based on the pre-trained network parameters, we update the three
networks by optimizing the overall loss with the selected hyper-parameters $\xi=0.5,\lambda=1.5,$ and $\eta=3$.
In addition, we also follow the same data augmentation strategy.
All the networks are trained for 40 epochs, using ADAM optimizer
with learning rate of 0.00035, momentum of 0.0005 and batch size
of 64. In each epoch, we use DBSCAN for clustering. It is note that only one model is kept during test.

    \begin{algorithm}
    \caption{SSKD's main training process.}
        \begin{algorithmic}[1] 
            \Require Source domain data: $\mathcal{S}=\{(x_{s,i}, y_{s,i})|_{i=1}^{n_{s}}\}$, target domain data: $\mathcal{T}=\{(x_{t,i})|_{i=1}^{n_{t}}\}$, batch size $B$.
            \Ensure The powerful model parameter.
            \State
            Initialize the parameters of encoder networks by optimizing with Eq. (1) on $\mathcal{S}$.
            \State 
            Draw three augmentation functions $\psi_{j}\sim \Psi, j=1,2,3.$
            \For{sampled batch $\{x_{i}\}_{i=1}^{B}}$ 
                \For{$i\in\{1,\cdots,B\}$}
                \State
                \# the augumentation
                \State
                $\tilde{x}_{t,3i+j}=\psi_{j}(x_{t,3i+j}), j=1,2,3.$
                \State
                \# representation
                \State
                $h_{t,3i+j}=f_{j}(\tilde{x}_{t,3i+j}), j=1,2,3.$
                \State
                \# projection
                \State
                $z_{t,3i+j}=g_{j}(h_{t,3i+j}), j=1,2,3.$
                \EndFor
                \State
                Update peer models by optimizing the objective function Eq. $(10)$.
                \State
                Update peer moving average models by Eq. $(5)$.
            \EndFor
            \State
            Return the powerful model parameters.
        \end{algorithmic}
    \end{algorithm}


\textbf{Momentum update mechanism}. The motivation behind
SSKD is that we generate self-supervised labels for one student model with
the prediction outputs from another network model, leading to transferring rich structured
knowledge between different models. Invariance learning has a important effect in this training process by using the momentum update mechanism. To verify its effect, we use one
network's current-iteration predictions as pseudo labels and remove the nuclear-norm maximization of batch classification response matrix in the slowly evolving networks. Such experiments
are denoted as SSKD (w/o $\theta_{\cdot,e}$). The large margin of performance declines (\emph{e.g.}, from 78.7\% to 46.3\% for mAP) can easily be observed in Table \ref{tab:Ablation_s}, which
demonstrates the importance of adopting the momentum update mechanism.

\begin{table*}[htbp]
  \centering
  \caption{Ablation studies of our proposed approach on individual components for Duke-to-Market and Market-to-Duke adaptation tasks.}
    \begin{tabular}{ccccc|cccc}
    \toprule
    \multicolumn{1}{c|}{\multirow{2}[4]{*}{Methods}} & \multicolumn{4}{c|}{Duke $\rightarrow$ Market} & \multicolumn{4}{c}{Market $\rightarrow$ Duke} \\
\cmidrule{2-9}    \multicolumn{1}{c|}{} & \multicolumn{1}{c|}{mAP} & \multicolumn{1}{c|}{top-1} & \multicolumn{1}{c|}{top-5} & top-10 & \multicolumn{1}{c|}{mAP} & \multicolumn{1}{c|}{top-1} & \multicolumn{1}{c|}{top-5} & top-10 \\
    \midrule
    \midrule
    \multicolumn{1}{l|}{SSKD (w/o $\mathcal{L}_{id}$)}  & 51.3  & 74.5  & 85.7  & 89.3  & 50.2  & 67.1    & 77.4  & 80.6 \\
    \multicolumn{1}{l|}{SSKD (w/o $\mathcal{L}_{dt}$)}  & 77.4  & 91.2  & 96.3  & 97.7  & 65.7    & 78.6  & 89.0  & 91.7 \\
    \multicolumn{1}{l|}{SSKD (w/o $\mathcal{L}_{BNM}$)}  & 77.8  & 90.6  & 96.8  & 98.0  & 66.4  & 79.7& 89.7  & 92.6 \\
    \multicolumn{1}{l|}{SSKD (w/o $\mathcal{L}_{MS}$)}  & 75.9  & 89.8  & 96.2  & 97.7    & 64.0  & 77.5  & 88.2    & 92.1 \\
    \multicolumn{1}{l|}{SSKD (w/o $\theta_{\cdot,e}$)}  & 46.3  & 76.0  & 86.9  & 90.5  & 42.5  & 63.5  & 76.0  & 80.0 \\
    \midrule
    \multicolumn{1}{c|}{SSKD} & 78.7 & 91.7 & 97.2 & 98.2 & 67.2 & 80.2 & 90.6 & 93.3 \\
    \bottomrule
    \end{tabular}%
  \label{tab:Ablation_s}%
\end{table*}%

\bibliographystyle{aaai}
\bibliography{refrence}

\end{document}